\documentclass[10pt,twocolumn,letterpaper]{article}

\usepackage{iccv}
\usepackage{times}
\usepackage{epsfig}
\usepackage{graphicx}
\usepackage{amsmath}
\usepackage{amssymb}

\usepackage{multirow}
\usepackage{booktabs}
\usepackage{caption}
\usepackage{graphics,color}

\usepackage[pagebackref=true,breaklinks=true,letterpaper=true,colorlinks,bookmarks=false]{hyperref}

\iccvfinalcopy 

\def\iccvPaperID{8645} 

\ificcvfinal\fi

\begin{document}
\twocolumn[{
\begin{@twocolumnfalse}

\title{Generating Smooth Pose Sequences for Diverse Human Motion Prediction}

\author{
Wei Mao$^1$, \;\;Miaomiao Liu$^{1}$,\;\; Mathieu Salzmann$^{2,3}$\;\; \\
$^1$Australian National University; $^2$CVLab, EPFL; $^3$ClearSpace, Switzerland\\
{\tt\small \{wei.mao, miaomiao.liu\}@anu.edu.au,}\;\;{\tt\small mathieu.salzmann@epfl.ch}
}

\maketitle
\ificcvfinal\thispagestyle{empty}\fi

\begin{center}
\setlength\tabcolsep{1pt}
\vspace{-0.8cm}\begin{tabular}{c}
      \includegraphics[width=\linewidth]{opening.pdf}
\end{tabular}
\end{center}
\vspace{-0.8cm}

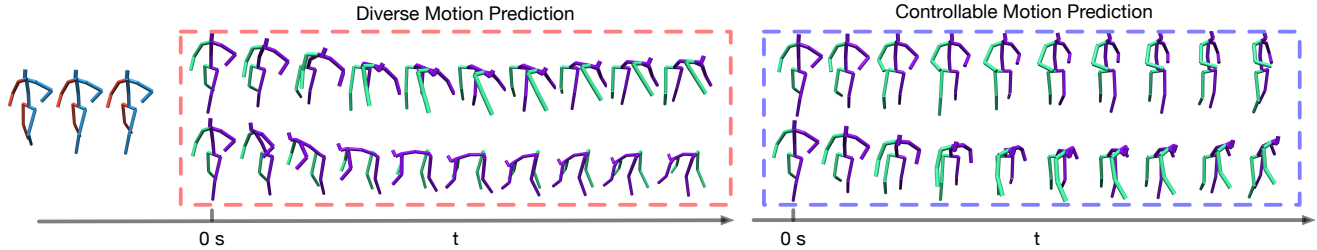
\captionof{figure}{{\bf Diverse and controllable motion prediction.} Given a past human motion, shown with blue and red skeletons, our model can predict diverse future motions, as shown in the red box. It also allows us to produce future motions with the same lower-body motion but diverse upper-body ones, i.e., to perform controllable motion prediction, as shown in the blue box. More qualitative results can be accessed at \url{https://youtu.be/mXpdmsEqF94}.
\vspace{0.2cm} }\label{fig:startfig}
\end{@twocolumnfalse}
}]

\begin{abstract}
    Recent progress in stochastic motion prediction, i.e., predicting multiple possible future human motions given a single past pose sequence, has led to producing truly diverse future motions and even providing control over the motion of some body parts. However, to achieve this, the state-of-the-art method requires learning several mappings for diversity and a dedicated model for controllable motion prediction. In this paper, we introduce a unified deep generative network for both diverse and controllable motion prediction. To  this end, we leverage the intuition that realistic human motions consist of smooth sequences of valid poses, and that, given limited data, learning a pose prior is much more tractable than a motion one. We therefore design a generator that predicts the motion of different body parts sequentially, and introduce a normalizing flow based pose prior, together with a joint angle loss, to achieve motion realism.
    Our experiments on two standard benchmark datasets, Human3.6M and HumanEva-I, demonstrate that our  approach outperforms the state-of-the-art baselines in terms of both sample diversity and accuracy. The code is available at \url{https://github.com/wei-mao-2019/gsps}  
\end{abstract}
\vspace{-0.5cm}
\section{Introduction}

Predicting future human motions from historical pose sequences has broad applications in autonomous driving~\cite{paden2016survey}, animation creation in the game industry~\cite{van2010real} and human robot interaction~\cite{koppula2013anticipating}. Most existing work focuses on deterministic prediction, namely, predicting only the most likely future sequence~\cite{fragkiadaki2015recurrent,Martinez_2017_CVPR,Li_2018_CVPR,mao2019learning, mao2020history}. However, future human motion is naturally diverse, especially over a long-term horizon ($>1$s). 

Most of the few attempts to produce diverse future motion predictions exploit variational autoencoders (VAEs) to model the multi-modal data distribution~\cite{walker2017pose,yan2018mt,aliakbarian2020stochastic}. 
These VAEs-based models are trained to maximize the motion likelihood.
As a consequence, and as discussed in~\cite{yuan2020dlow}, because training data cannot cover all possible diverse motions,
test-time sampling tends to concentrate on the major data distribution modes, ignoring the minor ones, and thus limiting the diversity of the output. 
To address this Yuan \etal~\cite{yuan2020dlow} proposed to learn multiple mapping functions, which  
produce multiple predictions that are explicitly encouraged to be diverse. 
While this framework indeed yields high diversity, it requires training several mappings in parallel, and separates the training of such mappings from that of the VAE it employs for prediction. Furthermore, while the approach was shown to be applicable to controllable motion prediction, doing so requires training a dedicated model and does not guarantee the controlled part of the motion, e.g., the lower body, to be truly fixed to the same motion in different predictions as the remaining body parts vary.

\begin{figure}[!ht]
    \centering
    \begin{tabular}{cc}
    \includegraphics[width=0.6\linewidth]{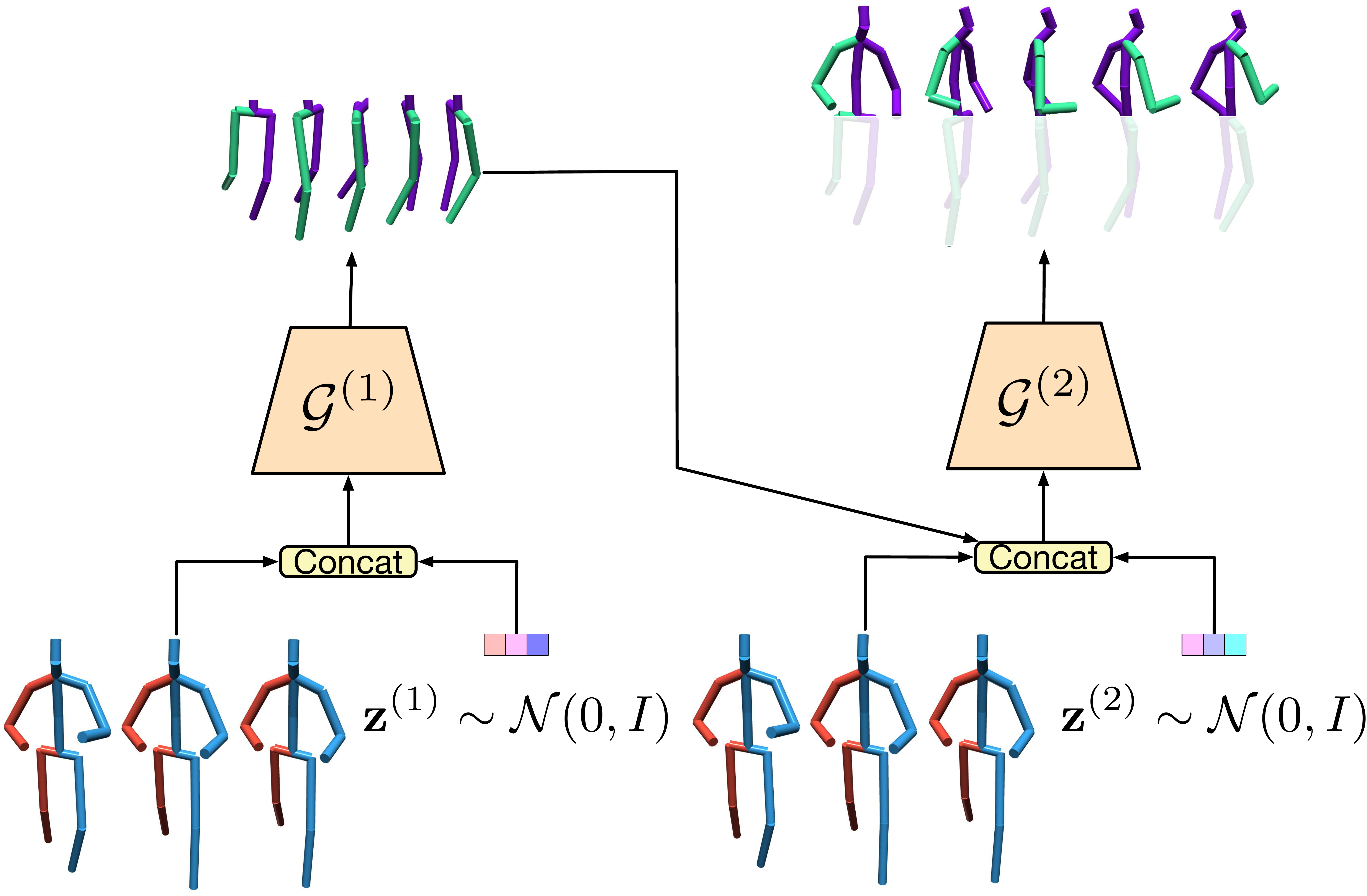} & 
    \includegraphics[width=0.25\linewidth]{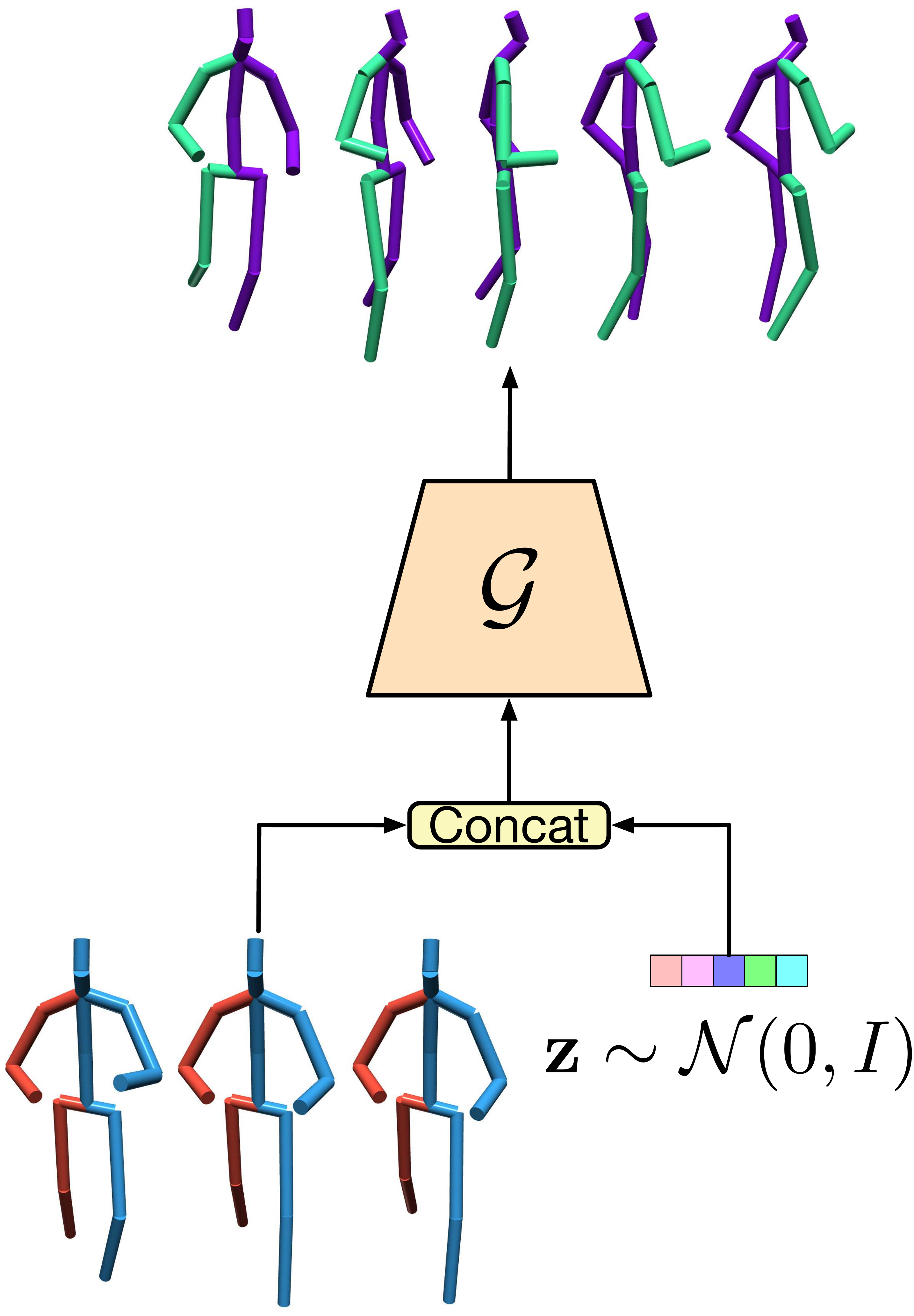}\\
    (a)&(b)
    \end{tabular}
    \vspace{-0.45cm}
    \caption{{\bf Our generator (a) vs a standard one (b).} (a) We predict the motions of different body parts sequentially.
    (b) Existing methods directly produce the whole motion.}
    \label{fig:generator}
    \vspace{-0.55cm}
\end{figure}
In this paper, we introduce an end-to-end trainable approach for diverse motion prediction that does not require learning several mappings to achieve diversity. Our framework yields fully controllable motion prediction; one can strictly fix the motion of one portion of the human body and generate diverse predictions for the other portion only.

To this end, we rely on the observation that diverse future motions are composed of valid human poses organized in smooth sequences. Therefore, instead of learning a \emph{motion} prior/distribution, for which sufficiently diverse training data is hard to obtain, we propose to learn a \emph{pose} prior and enforce a hard constraint on the predicted poses to form smooth sequences and to satisfy human kinematic constraints. Specifically, we model our pose prior as a normalizing flow~\cite{rezende2015variational,DinhSB17}, which allows us to compute the data log-likelihood exactly, and further promote diversity by maximizing the distance between pairs of samples during training.

To achieve controllable motion predictions, as illustrated in Fig.~\ref{fig:generator}~(a),  we generate the future motions of the different body parts of interest in a sequential manner. Our design allows us to produce diverse future motions that share the same partial body motion, e.g., the same leg motion but diverse upper-body motions. This is achieved by fixing the latent codes for some body parts while varying those of the other body parts. In contrast to~\cite{yuan2020dlow}, our approach allows us to train a single model that achieves both non-controllable and controllable motion prediction.

Our contributions can be summarized as follows: 
(i) We develop a unified framework achieving both diverse and part-based
controllable human motion prediction, using a pre-ordered part sequence;
(ii) We propose a \emph{pose} prior and a \emph{joint~angle} constraint to regularize the training of our generator and encourage it to produce smooth pose sequences. Such strategy overcomes the difficulty of learning the distribution of diverse \emph{motions} as other VAE-based methods do. 

Our experiments on standard human motion prediction benchmarks demonstrate that our approach outperforms the state-of-the-art methods in terms of sample diversity and accuracy. 
\begin{figure*}[!ht]
    \vspace{-0.25cm}
    \centering
    \includegraphics[width=\linewidth]{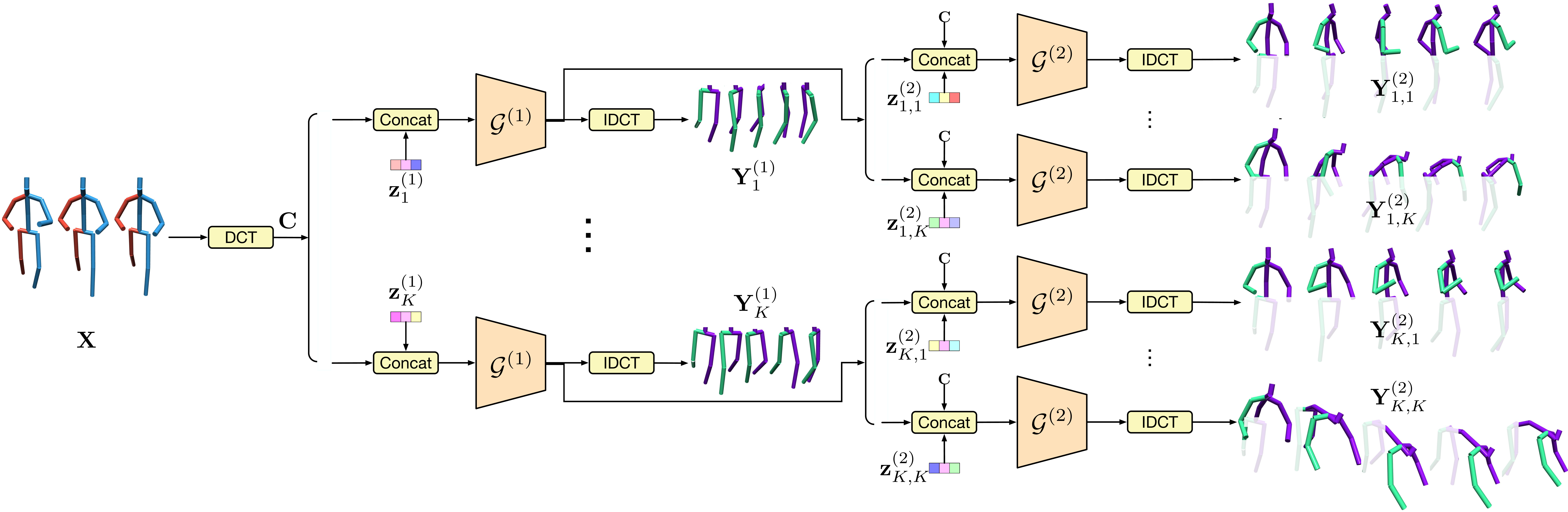} 
    \vspace{-0.8cm}
    \caption{{\bf Overview of our approach.} Given a past motion $\mathbf{X}$, we first sample $K$ latent codes $\{\mathbf{z}_{i}^{(1)}\}_{i=1}^{K}$ and decode them to future lower-body motions $\{\mathbf{Y}_{i}^{(1)}\}_{i=1}^{K}$. For each such future motion $\mathbf{Y}_{i}^{(1)}$, we again sample $K$ latent codes $\{\mathbf{z}_{i}^{(2)}\}_{i=1}^{K}$ to generate  future upper-body motions $\{\mathbf{Y}_{i,j}^{(2)}\}_{j=1}^{K}$.}
    \label{fig:pipeline}
    \vspace{-0.5cm}
\end{figure*}

\section{Related Work}
\noindent\textbf{Human motion prediction.} 
Early attempts at human motion prediction~\cite{brand2000style,sidenbladh2002implicit,taylor2007modeling,wang2008gaussian} relied on non-deep learning approaches, such as Hidden Markov Model~\cite{brand2000style} and Gaussian Process latent variable models~\cite{wang2008gaussian}. Despite their success on modeling simple periodic motions, more complicated ones are typically better handled via deep neural networks. Deep learning based methods can be roughly categorized into deterministic approaches 
and stochastic ones. 

Deterministic models focus
on predicting the most likely human future motion sequence given historical observations~\cite{fragkiadaki2015recurrent,JainZSS16,Butepage_2017_CVPR,Martinez_2017_CVPR,pavllo_quaternet_2018,gui2018adversarial,Li_2018_CVPR,aksan2019structured,mao2019learning,wang2019imitation,gopalakrishnan2019neural,mao2020history,cai2020learning,mao2021multi}. 
Motivated by the success of RNNs for sequence modeling~\cite{sutskever2011generating,kiros2015skip}, many such methods employ recurrent architectures~\cite{fragkiadaki2015recurrent,JainZSS16,Martinez_2017_CVPR,pavllo_quaternet_2018,gui2018adversarial,gopalakrishnan2019neural}. 
Nevertheless, feed-forward models have more recently been shown to effectively leverage the human kinematic structure and long motion history~\cite{Li_2018_CVPR,mao2019learning,aksan2019structured,Butepage_2017_CVPR}.
In any event, while deterministic human motion prediction models have achieved promising results, especially for short-term prediction ($<0.5$s), they struggle with predictions for long-term horizons ($>1$s). This is because human motion is an inherently stochastic process, where one observed motion can lead to multiple possible future ones. 

Addressing this has been the focus of stochastic motion prediction methods. Existing ones are mainly based on deep generative models~\cite{walker2017pose,lin2018human,barsoum2018hp,hernandez2019human,kundu2019bihmp,yan2018mt,aliakbarian2020stochastic,yuan2020dlow}, such as variational autoencoders (VAEs)~\cite{kingma2013auto} and generative adversarial networks (GANs)~\cite{goodfellow2014generative}. In the context of VAE-based ones, Yan~\etal~\cite{yan2018mt} proposed 
to jointly learn a feature embedding for motion reconstruction and a feature transformation to model the motion mode transitions; Aliakbarian~\etal~\cite{aliakbarian2020stochastic} introduced a perturbation strategy for the random variable to prevent the generator from ignoring it. In both cases, once the generator is trained, possible future motions are obtained by feeding it randomly-sampled latent codes. However, as argued in~\cite{yuan2020dlow}, such likelihood-based sampling strategy concentrates on the major mode(s) of the data distribution while ignoring the minor ones. 
To address this, Yuan \etal~\cite{yuan2020dlow} introduced a learnable sampling strategy equipped with a prior explicitly encouraging the diversity of the future predictions obtained from a \emph{pre-trained} generator.
Despite their promising performance, their future predictions are constrained by the use of a {\it pre-trained} generator. As an alternative to VAE-based models, GAN-based methods~\cite{lin2018human,barsoum2018hp,hernandez2019human,kundu2019bihmp} train the generator jointly with a discriminator. While, in principle, one could also employ a diversity-promoting prior in these methods, in practice, the resulting additional constraints further complicate the inherently-difficult training process~\cite{ArjovskyB17}. As such, existing GAN-based methods tend to produce limited diversity. 
Here, instead of using a discriminator to regularize the generation process, we employ a normalizing-flow-based \emph{pose} prior, accounting for the fact that training data can more easily cover the diversity of poses than that of motions, and encourage the resulting poses to be valid and form smooth sequences.

\noindent\textbf{3D human pose prior.} In the 3D human pose estimation literature, many works~\cite{bogo2016keep,pavlakos2019expressive,zanfir2020weakly} have attempted to learn 3D pose priors to avoid invalid human poses. Such priors include Gaussian mixture models~\cite{bogo2016keep} and VAEs~\cite{pavlakos2019expressive}. However, as discussed in~\cite{zanfir2020weakly}, these priors only approximate the pose log-likelihood and may lead to instability~\cite{zanfir2020weakly}. To evaluate the exact log-likelihood, Zanfir~\etal~\cite{zanfir2020weakly} thus proposed a normalizing flow based pose prior. Normalizing flows (NFs)~\cite{rezende2015variational,DinhSB17} have recently become popular for density estimation precisely because they allow one to compute the exact log-likelihood. Here, we use a pre-trained NF model to evaluate the log-likelihood of the generated future poses, and by maximizing the log-likelihood, encourage the generator to produce realistic poses.

\noindent\textbf{Joint angle constraint.} Joint angle limits have been explored for the task of human pose estimation~\cite{herda2005hierarchical,chen2013data,akhter2015pose,dabral2018learning}. In particular, Akhter~\etal~\cite{akhter2015pose} introduced a pose-dependent joint limit function learned from a motion capture dataset. However, their function is non-differentiable, which makes it ill-suited for deep neural networks. Our angle loss is similar to that of~\cite{dabral2018learning}. However, unlike Dabral~\etal~\cite{dabral2018learning} who only consider angle constraints on arms and legs, consisting of manually-defined valid angle ranges, we encompass all angles between different body parts/joints and exploit the training data to determine the valid angle ranges.

\noindent\textbf{Controllable motion prediction.} To the best of our knowledge, Dlow~\cite{yuan2020dlow} constitutes the only attempt at controllable motion prediction, e.g, fixed lower-body motion but diverse upper-body motions. This was achieved by training a dedicated model, different from the uncontrollable one, yet sill not guaranteeing absolute control of the body parts that should undergo the same motion.  
By contrast, we develop a unified model that can achieve both controllable and uncontrolled diverse  motion prediction, while guaranteeing the controlled parts to truly follow a fixed motion.

Controllable motion prediction/generation has also been studied in computer graphics, specifically for virtual character 
control~\cite{holden2016deep,holden2017phase,ling2020character}. These works focus on generating human motions
for a specific goal, such as following a given path or performing pre-defined actions. In particular, \cite{ling2020character} relies on a motion VAE to capture motion dynamics and searches for motions that achieve the desired task via sampling policies, while accounting for neither motion diversity nor the detailed body movements. By contrast, our work aims to predict diverse future motions and control the detailed motion of body parts.

\section{Our Approach}
Let us now introduce our approach to diverse and controllable motion prediction.
We represent a human pose $\mathbf{x}\in \mathbb{R}^D$ as the concatenation of 3D joint coordinates in a single frame. Given a past motion sequence $\mathbf{X}=[\mathbf{x}_1,\mathbf{x}_2,\cdots,\mathbf{x}_H]^T$ of $H$ frames, we aim to predict a set of  poses $\mathbf{Y} =[\mathbf{x}_{H+1},\mathbf{x}_{H+2},\cdots,\mathbf{x}_{H+T}]^T$ representing a possible future motion. To this end, we rely on a deep generative model, which we design to yield diversity and allow for controllable motion prediction, as discussed below.

\begin{figure*}[!t]
    \vspace{-0.5cm}
    \centering
    \begin{tabular}{cc}
    \includegraphics[width=0.43\linewidth]{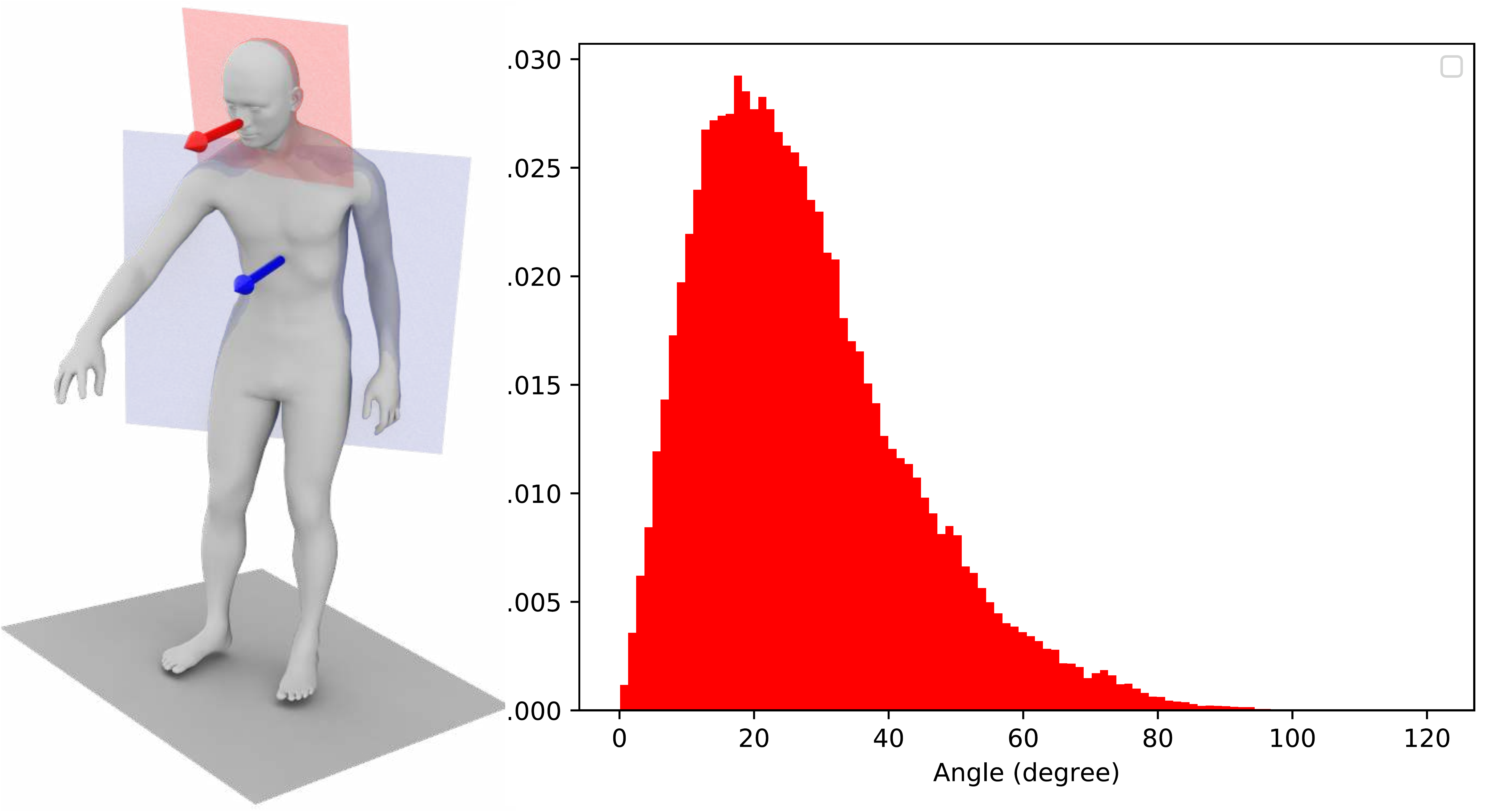} & 
    \includegraphics[width=0.43\linewidth]{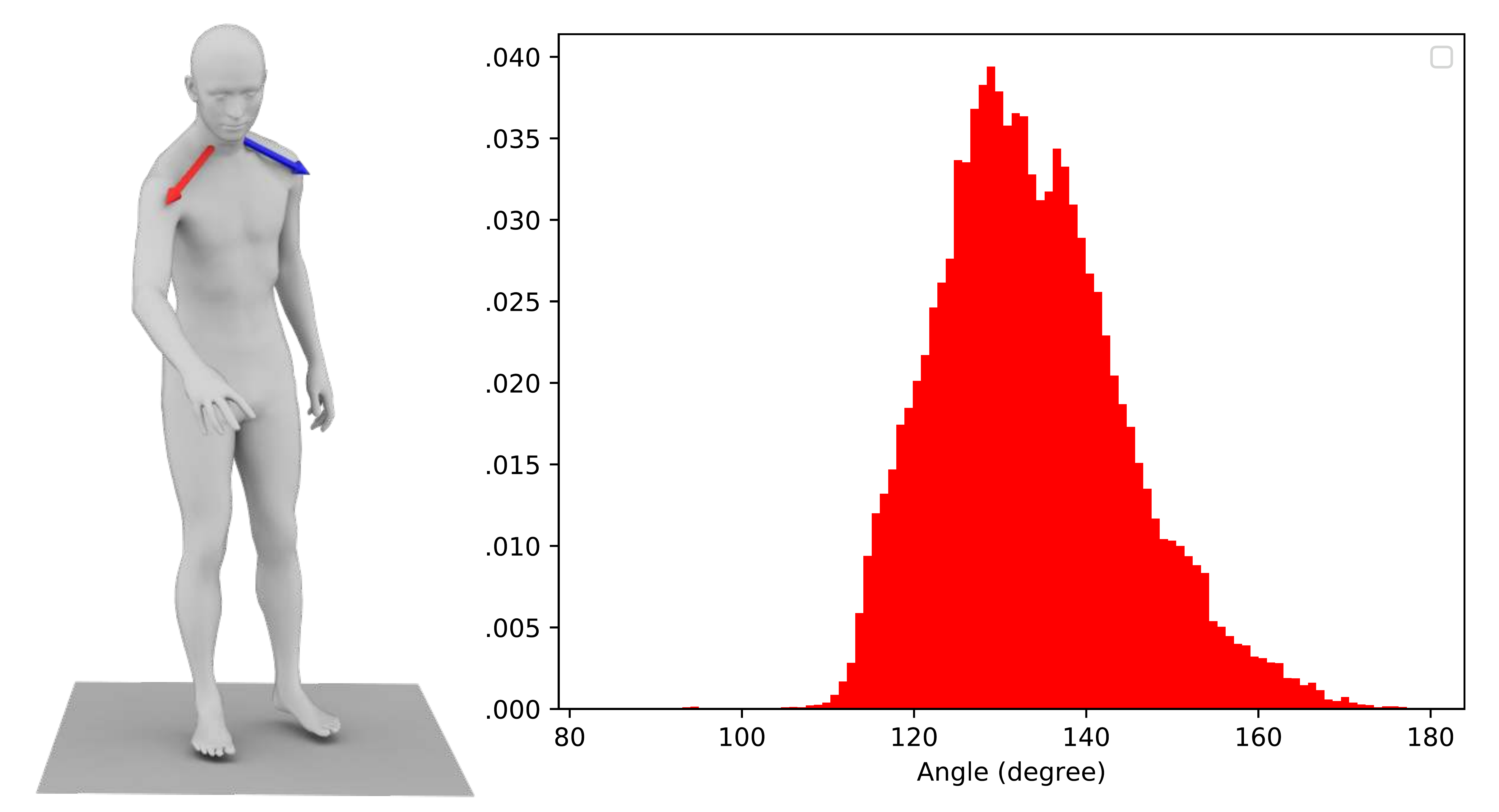}\\
    (a)&(b)
    \end{tabular}
    \vspace{-0.4cm}
    \caption{{\bf Examples of angle limits.} We consider the angles between the red and blue arrows on the left figures, and show their distributions in the right plots. We define the valid scope based on the minimal and maximal angles. (a) Angle between the head and body; (b) Angle between the left and right shoulders.}
    \label{fig:ang_lim}
    \vspace{-0.4cm}
\end{figure*}
\subsection{Diverse Motion Prediction}
In this section, we first briefly review the use of deep generative models in the context of human motion prediction and then introduce our solution for diverse prediction.

\noindent{\textbf{Deep Generative Models.}} 
Let $p(\mathbf{Y}|\mathbf{X})$ denote the data distribution of future motions $\mathbf{Y}\in \mathcal{Y}$, where $\mathcal{Y}$ is the set of all possible future motions, conditioned on past motions $\mathbf{X}\in \mathcal{X}$, with $\mathcal{X}$ encompassing all possible motion history.
By introducing a latent variable $\mathbf{z} \in \mathcal{Z}$, the data distribution can be reparameterized as $p(\mathbf{Y}|\mathbf{X}) = \int p(\mathbf{Y}|\mathbf{X}, \mathbf{z}) p(\mathbf{z})d\mathbf{z}$, where $p(\mathbf{z})$ is a Gaussian distribution. Generating a future motion $\mathbf{Y}$ can then be achieved by sampling a latent variable $\mathbf{z}$ and mapping it to $\mathbf{Y}$ using a deterministic \emph{generator} function $\mathcal{G}: \mathcal{Z}\times \mathcal{X} \rightarrow \mathcal{Y}$. Formally, this is expressed as
\begin{align}
\label{eq:pz}
&\mathbf{z} \sim p(\mathbf{z})\,, \\
\label{eq:gen}
&\mathbf{Y} = \mathcal{G}(\mathbf{z}, \mathbf{X})\,,
\end{align}
where the generator $\mathcal{G}$ is commonly implemented as a deep neural network. 
To train such a generator,
VAE-based methods~\cite{yan2018mt,aliakbarian2020stochastic} typically rely on maximizing the evidence lower bound~(ELBO) of the motion likelihood, i.e., minimizing the data reconstruction error and the KL
divergence.
However, this does not encourage diversity across different randomly-sampled vectors ${\bf z}$; instead it focuses on maximizing the likelihood of the training motions, and test-time diversity is thus limited by that of the training motions.

To overcome this, we propose to explicitly generate $K$ future motions $\{\Hat{\mathbf{Y}}_j\}_{j=1}^{K}$ for each sample during training, and explicitly promote their diversity. Because we aim for diversity, we cannot encourage all generated motions to match the ground truth. Therefore, we redefine the reconstruction error so that at least one of them is close to the ground truth. This yields the loss
    \vspace{-0.2cm}
\begin{equation}
    \mathcal{L}_{r} = \min_{j} \|\Hat{\mathbf{Y}}_j-\mathbf{Y}\|^2\;,
    \vspace{-0.2cm}
\end{equation}
where $j\in\{1,2,\cdots,K\}$ indexes over the generated motions for one sample. 

This loss, however, imposes constraints on only one of the $K$ generated motions. To better constrain the others, we leverage the intuition that, while the dataset contains only one ground-truth future motion for each past motion, several sequences have similar past motions. For each past motion, we therefore search for the training samples with similar past motions, based on a distance threshold, and take their future motion as pseudo ground truth. 
Let $\{\mathbf{Y}_{p}\}_{p=1}^{P}$ denote the resulting pseudo ground truths. Then, we define the multi-modal reconstruction error
\vspace{-0.2cm}
\begin{equation}
    \mathcal{L}_{mm} = \frac{1}{P}\sum_{p=1}^{P}\min_{j} \|\Hat{\mathbf{Y}}_j-\mathbf{Y}_p\|^2\;,
\vspace{-0.2cm}
\end{equation}
where $j\in\{1,2,\cdots,K\}$. It encourages our generator to cover each pseudo ground truth with at least one sampled motion.

To further explicitly encourage diversity across the generated motions, following~\cite{yuan2020dlow}, we  use the diversity-promoting loss
\vspace{-0.2cm}
\begin{equation}
    \mathcal{L}_{d} = \frac{2}{K(K-1)} \sum_{j=1}^{K}\sum_{k=j+1}^{K} e^{-\frac{\|\Hat{\mathbf{Y}}_{j}-\Hat{\mathbf{Y}}_{k}\|_1}{\alpha}}\;,
\end{equation}
where $\alpha$ is a normalizing factor. 

One drawback of this diversity loss, however, is that it may lead the model to produce unrealistic and physically-invalid motions, particularly if $P<K$ above, which leaves some motions unregularized. 
The most straightforward way to overcome this would be to learn a motion prior.
This, however, would require an impractically large amount of training data. Instead, we therefore leverage the observation that natural motions are composed of valid human poses and thus introduce the pose prior and angle losses discussed below.

\noindent{\textbf{Pose prior.}} We model our pose prior using a normalizing flow~\cite{rezende2015variational,DinhSB17}, which is an invertible transformation that aims to transfer an unknown data distribution to a distribution with a tractable density function, e.g., a Gaussian distribution. 
In other words, we model the 3D human pose distribution $p(\mathbf{x})$ by learning a bijective and differentiable function$f(\cdot)$, which maps a pose sample $\mathbf{x}\sim p(\mathbf{x})$ to a latent representation $\mathbf{h} = f(\mathbf{x})$ following a standard Gaussian distribution, i.e., $\mathbf{h}\sim\mathcal{N}(0,\mathbf{I})$.

Following the normalizing flow literature~\cite{rezende2015variational,DinhSB17}, this allows us to compute the likelihood of a pose $\mathbf{x}$ as
\vspace{-0.2cm}
\begin{equation}
    \label{eq:change_var}
    p(\mathbf{x}) = g(\mathbf{h}) \left|\det\left(\frac{\partial f}{\partial\mathbf{x}}\right)\right|\;,
    \vspace{-0.2cm}
\end{equation}
where $g(\mathbf{h})=\mathcal{N}(\mathbf{h}|0,\mathbf{I})$ and $\det(\frac{\partial f}{\partial\mathbf{x}})$ is the determinant of the Jacobian matrix of $f(\cdot)$.

In practice, the function $f$ is modeled via a deep network. Considering model size and inference efficiency, we choose a simple network with only 3 fully-connected layers, which is in stark contrast with the much larger architectures in the recent normalizing flow literature~\cite{DinhSB17,NIPS2016_ddeebdee,papamakarios2017masked,huang2018neural}. To ensure that $f(\cdot)$ is invertible, we compute the weights of each layer via a QR decomposition and use monotonic activation functions. More details about our normalizing flow architecture are provided in the supplementary material. 

Given a dataset $\mathcal{D}$ of valid human poses, we learn the function $f$ by maximizing the log-likelihood of the samples in $\mathcal{D}$. This can be written as \vspace{-0.2cm}
\begin{align}
    f^{*} &= {\arg\max}_{f}\sum_{\mathbf{x}\in\mathcal{D}}{\log{p(\mathbf{x})}}\\
    &={\arg\max}_{f}\sum_{\mathbf{x}\in\mathcal{D}}{\log{g(\mathbf{h}})+\log{ \left|\det\left(\frac{\partial f}{\partial\mathbf{x}}\right)\right|}}\;.
\vspace{-0.2cm}
\end{align}

Given the trained function $f$, we then define a loss function to encourage our generator to produce valid poses. Specifically, we write it as the negative log-likelihood of a generated human pose $\Hat{\mathbf{x}}$, which yields \vspace{-0.2cm}
\begin{align}
    \mathcal{L}_{nf} & = -\log p(\Hat{\mathbf{x}})\\
    &= -\log g(\Hat{\mathbf{h}}) - \log\left|\det\left(\frac{\partial f^{*} }{\partial\Hat{\mathbf{x}}}\right)\right|\;, 
    \vspace{-0.2cm}
\end{align}
where $\Hat{\mathbf{h}}=f(\Hat{\mathbf{x}})$ and $g(\Hat{\mathbf{h}})=\mathcal{N}(\Hat{\mathbf{h}}|0,\mathbf{I})$.

\noindent{\textbf{Joint angle loss.}} In addition to our learnt pose prior, we leverage the fact that human movements are constrained by the physiological structure of the human body, e.g., one cannot turn their head fully backwards. In our context, this means that the angles between some body parts are limited to a certain range. Here, instead of manually encoding these different ranges, we discover them by analysing a valid human pose dataset $\mathcal{D}$.

To this end, as shown in Fig.~\ref{fig:ang_lim}, we first compute unit length vectors, which represent either the orientations of body parts or the directions of limbs. A body part orientation is the normal of the plane defined by 3 joints. For example, the torso plane, shown in blue in Fig.~\ref{fig:ang_lim}~(a), is defined by the left and right shoulders and the pelvis. A limb direction is defined by 2 joints of the limb. We then compute the angle between those vectors for every pose in $\mathcal{D}$ and define the valid range based on the minimal and maximal such angle. We list all angles and their valid ranges in the supplementary material.

Let $\{a_j\}_{j=1}^{L}$ denote the $L$ pre-defined angles, and $l_{a_j}$ and $u_{a_j}$ the lower bound and upper bound of angle $a_j$, respectively. Given a human pose $\Hat{\mathbf{x}}$, we write our joint angle loss for $a_j$ as \vspace{-0.2cm}
\begin{align}
    \mathcal{L}_{a_j} = \begin{cases}
    (a_j(\Hat{\mathbf{x}})-l_{a_j})^2,\, \text{if } a_j < l_{a_j}\\
    (a_j(\Hat{\mathbf{x}})-u_{a_j})^2,\, \text{if } a_j > u_{a_j}\\
    0, \, \text{otherwise}\\
\end{cases}
\vspace{-0.2cm}
\end{align}
where $a_j(\Hat{\mathbf{x}})$ is the angle value calculated from pose $\Hat{\mathbf{x}}$. We then combine the different angles in our final angle loss $\mathcal{L}_{a}=\sum_{j=1}^{L}\mathcal{L}_{a_j}$.

\begin{figure}[h]
    \centering
    \includegraphics[width=0.9\linewidth]{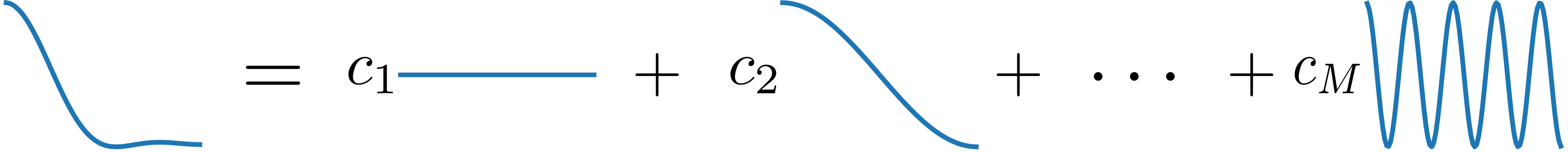}
    \vspace{-0.2cm}
    \caption{{\bf A smooth trajectory} can be compactly represented by a linear combination of predefined DCT bases~\cite{akhter2009nonrigid}.}
    \label{fig:dct_vis}
    \vspace{-0.2cm}
\end{figure}

\noindent{\textbf{Predicting smooth trajectory.}}
In the stochastic motion prediction literature, human movements are commonly represented by a sequence of 3D joint coordinates. With such a representation, encouraging the generated poses to be valid, as discussed above, does not ensure that the resulting sequence will look natural. To enforce temporal smoothness, we therefore adopt a trajectory representation based on the Discrete Cosine Transform (DCT), as suggested in~\cite{mao2019learning}. In particular, using a reduced number of low-frequency DCT components guarantees the resulting trajectory to be smooth as shown in Fig.~\ref{fig:dct_vis}. \vspace{-0.2cm}

Specifically, given the past motion sequence $\mathbf{X}$, we first replicate the last pose $T$ times to generate a temporal sequence $\Tilde{\mathbf{X}}$ of length $H+T$, where $T$ is the length of the future sequence to predict. We then compute the DCT coefficients of this sequence $\mathbf{C}$ as 
\vspace{-0.2cm}
\begin{equation}
    \mathbf{C} = \Tilde{\mathbf{X}}\mathbf{T}\;,
    \label{eq:dct} 
    \vspace{-0.2cm}
\end{equation}
where $\mathbf{T}\in\mathbb{R}^{(H+T)\times M}$ and each column of $\mathbf{T}$ represents a predefined DCT basis; $\Tilde{\mathbf{X}}\in \mathbb{R}^{D\times (H+T)}$ and each row of $\Tilde{\mathbf{X}}$ is the trajectory of a joint coordinate; $\mathbf{C}\in \mathbb{R}^{D\times M}$ and each row of $\mathbf{C}$ represents the first $M\leq H+T$ DCT coefficients for a trajectory.

We then make our generator predict the DCT coefficients of the future motion $\Hat{\mathbf{C}}$ given those of the past motion $\mathbf{C}$. Given the predicted coefficients, we recover the future motion via inverse DCT as 
\vspace{-0.2cm}
\begin{equation}
    \Hat{\mathbf{Y}} = \Hat{\mathbf{C}}\mathbf{T}^T\;,
    \label{eq:idct}
\vspace{-0.2cm}
\end{equation}
where $\Hat{\mathbf{C}}\in \mathbb{R}^{D\times M}$ and $\Hat{\mathbf{Y}}\in \mathbb{R}^{D\times (H+T)}$. As in~\cite{mao2019learning}, our generator also outputs the past motion to encourage the transition from past poses to future ones to be smooth. See the supplementary material for more detail.

\begin{table*}[!ht]
	\footnotesize
	\centering
	\resizebox{0.94\textwidth}{!}{
		\begin{tabular}{@{\hskip 0mm}cccccccccccc@{\hskip 0mm}}
			\toprule
			\multirow{2}{*}{Method}& \multicolumn{5}{c}{Human3.6M~\cite{h36m_pami}} & & \multicolumn{5}{c}{HumanEva-I~\cite{sigal2010humaneva}} \\ \cmidrule{2-6} \cmidrule{8-12}
			 & APD $\uparrow$ & ADE $\downarrow$ & FDE $\downarrow$ & MMADE $\downarrow$ & MMFDE $\downarrow$ & & APD $\uparrow$ & ADE $\downarrow$ & FDE $\downarrow$ & MMADE $\downarrow$ & MMFDE $\downarrow$ \\ \midrule
			ERD \cite{fragkiadaki2015recurrent}                     & 0  & 0.722 & 0.969 & 0.776 & 0.995 &  & 0 & 0.382 & 0.461 & 0.521 & 0.595 \\
			acLSTM \cite{zhou2018auto}                      & 0  & 0.789 & 1.126 & 0.849 & 1.139 &  & 0 & 0.429 & 0.541 & 0.530 & 0.608 \\\hline
			Pose-Knows \cite{walker2017pose}              & 6.723  & 0.461 & 0.560 & 0.522 & 0.569 &  & 2.308 & 0.269 & 0.296 & 0.384 & 0.375 \\
			MT-VAE \cite{yan2018mt}           & 0.403  & 0.457 & 0.595 & 0.716 & 0.883 &  & 0.021 & 0.345 & 0.403 & 0.518 & 0.577 \\
			HP-GAN \cite{barsoum2018hp}           & 7.214  & 0.858 & 0.867 & 0.847 & 0.858 &  & 1.139 & 0.772 & 0.749 & 0.776 & 0.769 \\\hline
			BoM \cite{bhattacharyya2018accurate} & 6.265  & 0.448 & 0.533 & 0.514 & 0.544 &  & 2.846 & 0.271 & 0.279 & 0.373 & 0.351 \\
			GMVAE \cite{dilokthanakul2016deep}            & 6.769  & 0.461 & 0.555 & 0.524 & 0.566 &  & 2.443 & 0.305 & 0.345 & 0.408 & 0.410 \\
			DeLiGAN \cite{gurumurthy2017deligan}          & 6.509  & 0.483 & 0.534 & 0.520 & 0.545 &  & 2.177 & 0.306 & 0.322 & 0.385 & 0.371 \\
			DSF \cite{yuan2019diverse}                    & 9.330  & 0.493 & 0.592 & 0.550 & 0.599 &  & 4.538 & 0.273 & 0.290 & 0.364 & 0.340 \\
			DLow~\cite{yuan2020dlow} & 11.741 & 0.425 & 0.518 & 0.495 & 0.531 &  & 4.855 & 0.251 & 0.268 & 0.362 & 0.339 \\\hline
			Ours & \textbf{14.757} & \textbf{0.389} & \textbf{0.496} & \textbf{0.476} & \textbf{0.525} & &\textbf{5.825} & \textbf{0.233} & \textbf{0.244} & \textbf{0.343} & \textbf{0.331}\\
			\bottomrule
		\end{tabular}
	}
	\vspace{-0.3cm}
	\caption{\textbf{Quantitative results} on Human3.6M and HumanEva-I. Our model consistently outperforms others on all metrics.}
	\label{tab:quan}
	\vspace{-0.4cm}
\end{table*}
\subsection{Controllable Motion Prediction}
The diverse motion prediction framework discussed above does not provide any control over the generated motions.
For controllable motion prediction, our goal is to predict future sequences that share the same motion
for parts of the body while being diverse for the other parts, e.g., same leg motion but diverse upper-body motion. To this end, instead of directly modeling the joint data distribution as discussed above, we propose to model a product of sequential conditional distributions.

Specifically, let a human motion be split into $N$ different body part motions, that is, $\mathbf{Y}=[\mathbf{Y}^{(1)},\mathbf{Y}^{(2)},\cdots,\mathbf{Y}^{(N)}]$, where $\mathbf{Y}^{(i)}\in \mathbb{R}^{T\times D_i}$ defines the motion of the $i^{th}$ body part, e.g., the left leg. Then, the future body motion distribution $p(\mathbf{Y}|\mathbf{X})$ can be expressed as
\vspace{-0.2cm}
\begin{equation}
    \label{eq:pyx}
    \resizebox{0.87\linewidth}{!}{
    $p(\mathbf{Y}|\mathbf{X}) = p(\mathbf{Y}^{(1)}|\mathbf{X})p(\mathbf{Y}^{(2)}|\mathbf{X},\mathbf{Y}^{(1)})\cdots p(\mathbf{Y}^{(N)}|\mathbf{X},\{\mathbf{Y}^{(i)}\}_{i=1}^{N-1})$\;.}
\vspace{-0.2cm}
\end{equation}

Each of the $N$ conditional distributions describes the motion of a particular body part given the motion of the previous body parts. Similar to the standard deep generative model discussed above, we model each conditional distribution as
\vspace{-0.2cm}
\begin{equation}
\label{eq:pzi}
\mathbf{z}^{(i)} \sim p^{(i)}(\mathbf{z})\,,
\vspace{-0.2cm}
\end{equation}
\vspace{-0.2cm}
\begin{equation}
\label{eq:geni}
\mathbf{Y}^{(i)} = \mathcal{G}^{(i)}(\mathbf{z}^{(i)}, \mathbf{X},\{\mathbf{Y}^{(j)}\}_{j=1}^{i-1})\,,
\vspace{-0.2cm}
\end{equation}
where $i \in \{1,2,\cdots, N\}$ and $p^{(i)}(\mathbf{z})$ is a standard Gaussian distribution $\mathcal{N}(0,1)$. In other words, to compute a future motion, we sample different random variables $\{\mathbf{z}^{(i)}\}_{i=1}^N$ and decode them sequentially to obtain the future motion of each body part. By fixing some random variables $\{\mathbf{z}^{(i)}\}_{i=1}^J$ while varying the others $\{\mathbf{z}^{(i)}\}_{i=J+1}^N$, our model can generate multiple times the same motion for body parts $\{\mathbf{Y}^{(i)}\}_{i=1}^J$ while producing diverse motions for the other body parts $\{\mathbf{Y}^{(i)}\}_{i=J+1}^N$.

Note that our generator design lets us achieve both diverse and controllable motion prediction at the same time. During training, for each past motion of the $i$-th body part, we generate $K$ motions for the $(i+1)$-th body part leading to $K^N$ future motions given one past motion. 
For example, as illustrated in Fig.~\ref{fig:pipeline}, we sample $K$ different leg motions $\{\mathbf{Y}_{j}^{(1)}\}_{j=1}^{K}$ and for each leg motion $\mathbf{Y}_{j}^{(1)}$, we generate $K$ different upper-body motions $\{\mathbf{Y}_{j,k}^{(2)}\}_{k=1}^{K}$.
In this context, we then re-write our diversity-promoting loss as a per-body-part loss, leading to
\vspace{-0.2cm}
\begin{align}
    \mathcal{L}_{d_i} = \frac{2}{K(K-1)} \sum_{j=1}^{K}\sum_{k=j+1}^{K} e^{-\frac{\|\Hat{\mathbf{Y}}_{\cdot,j}^{(i)}-\Hat{\mathbf{Y}}_{\cdot,k}^{(i)}\|_1}{\alpha^{(i)}}}\;,
\vspace{-0.2cm}
\end{align}
where $i \in \{1,2,\cdots,N\}$ is the index of the body part, 
and $\mathbf{Y}_{\cdot, k}^{(i)}$ is the $k$-th future motion for the $i$-th body part\footnote{Note that we eliminate the sample index of previous body parts for simplicity. For example, the $k$-th future motion of the $2$nd body part should be represented as $\mathbf{Y}_{j,k}^{(2)}$, where $j\in \{1,2,\cdots, K\}$ is the sample index of the $1$st body part.}. Altogether, our final training loss is then expressed as
\vspace{-0.2cm}
\begin{align}
\resizebox{0.85\linewidth}{!}{
    $\mathcal{L} = \lambda_{nf}\mathcal{L}_{nf} + \lambda_{a}\mathcal{L}_{a} + \sum_{i=1}^{N}\lambda_{d_i}\mathcal{L}_{d_i} + \lambda_{r}\mathcal{L}_{r} + \lambda_{mm}\mathcal{L}_{mm}\;,$}
\vspace{-0.2cm}
\end{align}
where $N$ is the number of body parts.

In practice, inspired by~\cite{mao2019learning}, we define each generator $\mathcal{G}^{(i)}$ as a Graph Convolutional Network (GCN)~\cite{kipf2016semi} with several graph convolution layers. Given a feature map $\mathbf{F}\in \mathbb{R}^{D\times |\mathbf{F}|}$, each such layer computes a transformed feature map $\mathbf{F}^{'}\in \mathbf{R}^{D\times |\mathbf{F}^{'}|}$ as $\mathbf{F}^{'} =\tanh( \mathbf{A}\mathbf{F}\mathbf{W})$,
where $\mathbf{A}\in \mathbb{R}^{D\times D}$ represents a fully connected graph with learnable connectivity and $\mathbf{W}\in \mathbb{R}^{|\mathbf{F}|\times |\mathbf{F}^{'}|}$ is a matrix of trainable weights. The details of our network architecture are provided in the supplementary material.
\section{Experiments}
\subsection{Dataset}
Following~\cite{yuan2020dlow}, we evaluate our method on 2 motion capture datasets, Human3.6M~\cite{h36m_pami} and HumanEva-I~\cite{sigal2010humaneva}, and adopt the same training and testing settings as in~\cite{yuan2020dlow} on both datasets.

\noindent{\textbf{Human3.6M}} 
consists of 7 subjects performing 15 actions. We use 5 subjects (S1, S5, S6, S7, S8) for training and the remaining 2 subjects (S9, S11) for testing. We use the original frame rate (50 Hz) and a 17-joint skeleton. We remove the global translation. Our model is trained to observe 25 past frames (0.5s) and predict 100 future frames (2s).

\noindent{\textbf{HumanEva-I}} consists of 3 subjects performing 5 actions, depicted by videos captured at 60 Hz. A person is represented by a 15-joint skeleton. We adopt the official train/test split~\cite{sigal2010humaneva} and also remove the global translation. The model predicts 60 future frames (1s) given 15 past frames (0.25s).

\subsection{Metrics, Baselines \& Implementation}
\begin{figure*}[!t]
    \centering
    \includegraphics[width=0.92\linewidth]{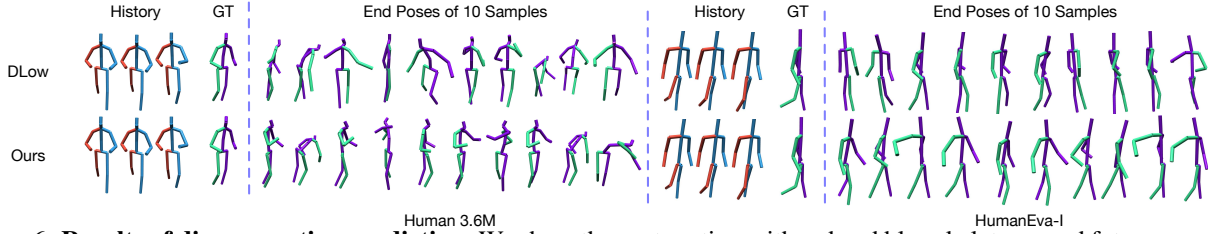}
    \vspace{-0.4cm}
    \caption{\textbf{Results of diverse motion prediction.} We show the past motion with red and blue skeletons, and future poses with green and purple ones.}
    \label{fig:results_vis}
    \vspace{-0.3cm}
\end{figure*}
\begin{figure*}[!ht]
    \centering
    \includegraphics[width=0.92\linewidth]{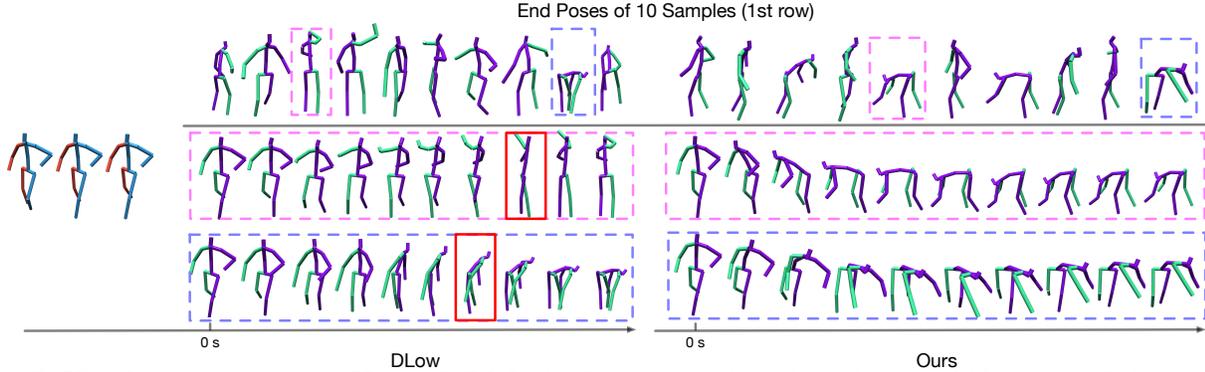}
    \vspace{-0.4cm}
    \caption{\textbf{Visualization of poses on Human3.6M.} In the first row, we show the end poses of 10 samples. Below, we then show different frames corresponding to the two samples highlighted by the magenta and blue dashed boxes. As highlighted by the red boxes, DLow~\cite{yuan2020dlow} yields invalid poses. This is because it lacks a pose level and kinematics prior.}
    \label{fig:h36m_vis}
    \vspace{-0.4cm}
\end{figure*}
\begin{figure*}[!ht]
    \centering
    \includegraphics[width=0.92\linewidth]{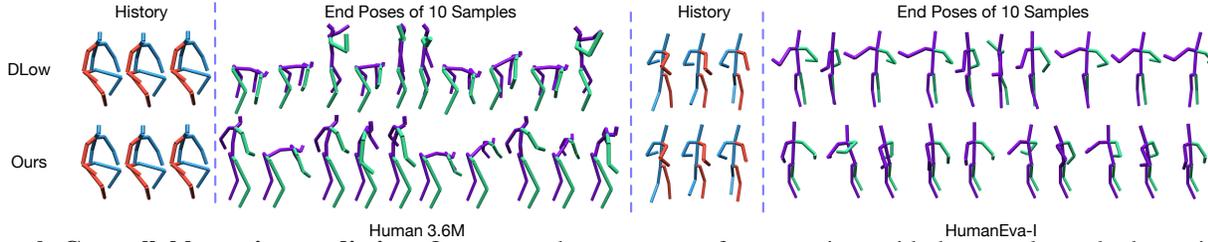}
    \vspace{-0.4cm}
    \caption{\textbf{Controllable motion prediction.} Our approach can generate future motions with the same lower-body motion but diverse upper-body ones.}
    \label{fig:results_vis_control}
    \vspace{-0.6cm}
\end{figure*}

\noindent{\textbf{Metrics.}} We follow the same evaluation protocol as in~\cite{yuan2020dlow} to measure diversity and accuracy. (1) To measure the prediction diversity, we use the Average Pairwise Distance (APD) defined as
$\frac{2}{K(K-1)}\sum_{i=1}^K \sum_{j=i+1}^K \|\Hat{\mathbf{Y}}_i - \Hat{\mathbf{Y}}_j\|_2$. (2) To measure the reconstruction accuracy over \emph{the whole sequence}, we use the Average Displacement Error (ADE) computed as $\frac{1}{T}\min\limits_{i} \|\Hat{\mathbf{Y}}_i - \mathbf{Y}\|_2$. (3) To measure the reconstruction accuracy of \emph{the last future pose}, we use the Final Displacement Error (FDE) defined as $\min\limits_{i} \|\Hat{\mathbf{Y}}_i[T] - \mathbf{Y}[T]\|_2$. We further report (4) the multi-modal version of ADE (MMADE), similar to $\mathcal{L}_{mm}$, and (5) the multi-modal version of FDE (MMFDE).

\begin{table}[!ht]
\footnotesize
\centering
\begin{tabular}{@{\hskip 0mm}cccccc@{\hskip 0mm}}
	\toprule
	\multirow{2}{*}{APD}& \multicolumn{2}{c}{Human3.6M~\cite{h36m_pami}} & & \multicolumn{2}{c}{HumanEva-I~\cite{sigal2010humaneva}} \\ \cmidrule{2-3} \cmidrule{5-6}
    &Lower $\downarrow$ & Upper $\uparrow$ & &Lower $\downarrow$ & Upper $\uparrow$\\ \midrule
    BoM~\cite{bhattacharyya2018accurate} w/ Eq.~\ref{eq:pyx} & 0 & 4.408 && 0 & 1.319 \\
    DLow~\cite{yuan2020dlow}-Control & 1.071 & 12.741 && 0.937 & 4.671 \\
    DLow~\cite{yuan2020dlow} w/ RS & 0.780 & 7.280 && 0.571 & 1.821\\\hline
    Ours & \textbf{0} & \textbf{13.150} && \textbf{0} & \textbf{5.096}\\
    \bottomrule
\end{tabular}
\vspace{-0.3cm}
\caption{APD for controllable motion prediction. From first to last row, we show results of BoM~\cite{bhattacharyya2018accurate} with our conditional formulation (Eq.~\ref{eq:pyx}), the controllable version of DLow~\cite{yuan2020dlow}, the controllable version of DLow~\cite{yuan2020dlow} with rejection sampling (RS) and our model respectively. We outperform DLow~\cite{yuan2020dlow} on both lower-body and upper-body. Note that DLow~\cite{yuan2020dlow} uses a different model here, whereas our results are obtained using the same model as in Table~\ref{tab:quan}.}
\label{tab:control}
\vspace{-0.6cm}
\end{table}
\noindent{\textbf{Baselines.}}
We compare our method with 3 types of baselines. (1) Deterministic motion prediction methods, including \textbf{ERD}~\cite{fragkiadaki2015recurrent} and \textbf{acLSTM}~\cite{zhou2018auto}; (2) Stochastic motion prediction methods without diversity-promoting technique, including CVAE based methods, \textbf{Pose-Knows}~\cite{walker2017pose} and \textbf{MT-VAE}~\cite{yan2018mt}, as well as a CGAN based method, \textbf{HP-GAN}~\cite{barsoum2018hp}; (3) Diverse motion prediction methods, including \textbf{BoM}~\cite{bhattacharyya2018accurate}, \textbf{GMVAE}~\cite{dilokthanakul2016deep}, \textbf{DeLiGAN}~\cite{gurumurthy2017deligan}, \textbf{DSF}~\cite{yuan2019diverse}, and \textbf{DLow}~\cite{yuan2020dlow}. The results of all baselines are directly reported from\cite{yuan2020dlow}.

\begin{table*}[!ht]
	\centering
	\resizebox{0.92\linewidth}{!}{
		\begin{tabular}{@{\hskip 0mm}cccccccccccccccccc@{\hskip 0mm}}
			\toprule
			\multirow{2}{*}{$\mathcal{L}_{nf}$}&\multirow{2}{*}{$\mathcal{L}_{a}$} & \multirow{2}{*}{$\mathcal{L}_{d}$} & \multirow{2}{*}{$\mathcal{L}_{r}$} & \multirow{2}{*}{$\mathcal{L}_{mm}$} & \multicolumn{6}{c}{Human3.6M~\cite{h36m_pami}} & & \multicolumn{6}{c}{HumanEva-I~\cite{sigal2010humaneva}} \\ 
			\cmidrule{6-11} \cmidrule{13-18}
			 &&&&& APD $\uparrow$ & ADE $\downarrow$ & FDE $\downarrow$ & MMADE $\downarrow$ & MMFDE $\downarrow$ & NLL $\downarrow$ & & APD $\uparrow$ & ADE $\downarrow$ & FDE $\downarrow$ & MMADE $\downarrow$ & MMFDE $\downarrow$ & NLL $\downarrow$ \\ \midrule
			 & \checkmark & \checkmark & \checkmark & \checkmark & 15.257 & 0.389 & 0.497 & 0.477 & 0.527 & 251.190 &  & \textbf{7.574} & 0.245 & 0.293 & 0.416 & 0.430 & 363.838 \\
			\checkmark &  & \checkmark & \checkmark & \checkmark & 19.608 & 0.397 & 0.534 & 0.506 & 0.575 & 97.774 &  & 6.569 & 0.235 & 0.276 & 0.406 & 0.410 & 119.673 \\
			\checkmark & \checkmark &  & \checkmark & \checkmark & 6.318 & \textbf{0.370} & \textbf{0.488} & 0.478 & 0.529 & \textbf{64.731} &  & 2.048 & \textbf{0.214} & 0.259 & 0.384 & 0.398 & \textbf{75.375}\\
			\checkmark & \checkmark & \checkmark &  & \checkmark & \textbf{20.030} & 0.479 & 0.562 & 0.513 & 0.569 & 89.726 &  & 6.778 & 0.567 & 0.625 & 0.606 & 0.633 & 109.777\\
			\checkmark & \checkmark & \checkmark & \checkmark &  & 18.079 & 0.394 & 0.538 & 0.520 & 0.587 & 91.977 &  & 6.474 & 0.234 & 0.283 & 0.415 & 0.421 & 113.630\\
			\checkmark & \checkmark & \checkmark & \checkmark & \checkmark & 14.757 & 0.389 & 0.496 & \textbf{0.476} & \textbf{0.525} & 74.872 & & 5.826 & 0.233 & \textbf{0.244} & \textbf{0.343} & \textbf{0.331} & 103.306\\
			\bottomrule
		\end{tabular}
	}
    \vspace{-0.25cm}
	\caption{\textbf{Ablation Study} on Human3.6M and HumanEva-I.}
	\label{tab:ablation}
\vspace{-0.7cm}
\end{table*}
\noindent{\textbf{Implementation.}} We set the hidden size of the generator $\mathcal{G}^{(i)}$ to 256 and the dimension of the random variable $\mathbf{z}^{(i)}$ to 64. To compare with the controllable version of DLow~\cite{yuan2020dlow}, we divide a human pose into 2 parts: lower and upper body ($N=2$). We also provide qualitative results for $N>2$ in our supplementary material. The number of samples $K$ is set to 10. Thus, during training, we predict 10 future lower body motions, and for each motion, we generate 10 upper body motions. For Human3.6M, the model is trained using a batch size of 16 for 500 epoch with 5000 training examples per epoch. The weights of the different loss terms $(\lambda_{nf},\lambda_{a},\lambda_{d_1},\lambda_{d_2},\lambda_{r},\lambda_{mm})$ and the normalizing factors $(\alpha_1,\alpha_2)$ are set to $(0.01,100,8,25,2,1)$ and $(100,300)$, respectively. For HumanEva-I, the model is trained using a batch size of 16 for 500 epoch with 2000 training examples per epoch. The weights of different loss terms $(\lambda_{nf},\lambda_{a},\lambda_{d_1},\lambda_{d_2},\lambda_{r},\lambda_{mm})$ and the normalizing factors $(\alpha_1,\alpha_2)$ are set to $(0.01,100,5,10,2,1)$ and $(15,50)$, respectively. Additional implementation details are provided in the supplementary material.

\subsection{Results}

\noindent{\textbf{Diverse Motion Prediction.}} 
In Table~\ref{tab:quan}, we compare our results with the baselines on Human3.6M and HumanEva-I. For all stochastic motion prediction baselines, the results are computed over 50 future motions per test sequence. 

Our method consistently outperforms all baselines on all metrics. In general, stochastic motion prediction methods outperform deterministic ones in accuracy (ADE, FDE, MMADE, MMFDE). The reason is that, for multi-modal datasets, deterministic prediction models tend to predict an average mode, which leads to higher errors. For stochastic motion prediction, there is a trade-off between sample diversity (APD) and accuracy. One can achieve high diversity while sacrificing some performance in accuracy, e.g., DSF~\cite{yuan2019diverse}, or vice versa, e.g., BoM~\cite{bhattacharyya2018accurate}.

Let us now focus on DLow~\cite{yuan2020dlow}, which constitutes the state of the art. Although, its learnable sampling strategy balances diversity and accuracy well, leading to better results than the other baselines, our method outperforms it on all metrics for both datasets. Note that, on Human3.6M, our method achieves $8\%$ lower reconstruction error (ADE) with $25\%$ higher sample diversity (APD). The qualitative comparisons in Fig.~\ref{fig:results_vis} further evidence that our predictions are closer to the GT and more diverse. We further provide a detailed comparison in Fig.~\ref{fig:h36m_vis}, which shows that DLow~\cite{yuan2020dlow} still produces some invalid poses, highlighted with red boxes. This can be caused by its lack of a pose-level prior.

\noindent{\textbf{Controllable Motion Prediction.}} 
We also compare our method with DLow for controllable motion prediction in Table~\ref{tab:control}. Here, the prediction model aims to predict future motions with the \emph{same} lower-body motion but \emph{diverse} upper-body motions. Our method gives a \emph{full} control of the lower-body with \emph{higher diversity} on upper-body. By contrast, DLow~\cite{yuan2020dlow} cannot guarantee the lower-body motion for different samples to be exactly the same. Although rejection sampling\footnote{For each test sequence, we sampled 1000 future motions and chose the 50 with lower-body motion closest to the target one.} helps to achieve a better control of the lower-body motion (third row), the diversity of upper body motion also drops. In Fig.~\ref{fig:results_vis_control}, we compare our results with those of DLow~\cite{yuan2020dlow}, which further supports our conclusions. Moreover, DLow~\cite{yuan2020dlow} requires a different model for controllable motion prediction, whereas our method yields a unified model able to achieve diverse and controllable motion prediction jointly. We further adapt our conditional formulation (Eq.~\ref{eq:pyx})
to one of the most recent baselines (BoM~\cite{bhattacharyya2018accurate}). The results in Table~\ref{tab:control} confirm that it also applies to other generators. 

\subsection{Ablation Study}
In Table~\ref{tab:ablation}, we evaluate the influence of our different loss terms. In general, there is a trade-off between the diversity loss $\mathcal{L}_{d}$ and the other losses. Without the diversity loss, the model yields the best ADE at the cost of diversity. By contrast, removing any other loss term leads to higher diversity but sacrifices performance on the corresponding accuracy metrics. Note that, in Table~\ref{tab:ablation}, we also report the negative log-likelihood (NLL) of the poses obtained from our pose prior to demonstrate their quality. Although, without the pose prior loss $\mathcal{L}_{nf}$ (first row), we can achieve higher diversity and almost the same accuracy, such diversity gain comes with a dramatic decrease in pose quality (NNL). In other words, while some samples are accurate, many others are not realistic. For qualitative comparison, please refer to the supplementary material.

\section{Conclusion}
In this paper, we have introduced an end-to-end trainable approach for both diverse and controllable motion prediction. To overcome the likelihood sampling problem that reduces sample diversity, we have developed a normalizing flow based pose prior together with a joint angle loss to encourage producing realistic poses, while enforcing temporal smoothness. To achieve controllable motion prediction, we have designed our generator to decode the motion of different body parts sequentially. Our experiments have demonstrated the effectiveness of our approach. Our current model assumes a predefined sequence of body parts, thus not allowing one to control an arbitrary part at test time. We will focus on addressing this in the future.

\noindent{\textbf{Acknowledgements}}

This research was supported in part by the Australia Research Council DECRA Fellowship (DE180100628) and ARC Discovery Grant (DP200102274). The authors would like to thank NVIDIA for the donated GPU (Titan V).

{\small
\bibliographystyle{ieee_fullname}
\bibliography{egbib.bib}
}

\pagebreak
\setcounter{equation}{0}
\setcounter{figure}{0}
\setcounter{table}{0}
\setcounter{page}{1}
\twocolumn[{
	\begin{@twocolumnfalse}
		\begin{center}
			\textbf{\Large Generating Smooth Pose Sequences for Diverse Human Motion Prediction \\
				-----Supplementary Material-----\\}
			
			\author{\; \\
				Wei Mao$^1$, \;\;Miaomiao Liu$^{1}$,\;\; Mathieu Salzmann$^{2,3}$\;\; \\
			$^1$Australian National University; $^2$CVLab, EPFL; $^3$ClearSpace, Switzerland\\
			{\tt\small \{wei.mao, miaomiao.liu\}@anu.edu.au,}\;\;{\tt\small mathieu.salzmann@epfl.ch}
		}
		\end{center}
	
	\begin{center}
		\setlength\tabcolsep{1pt}
		\begin{tabular}{c}
			\includegraphics[width=0.85\linewidth]{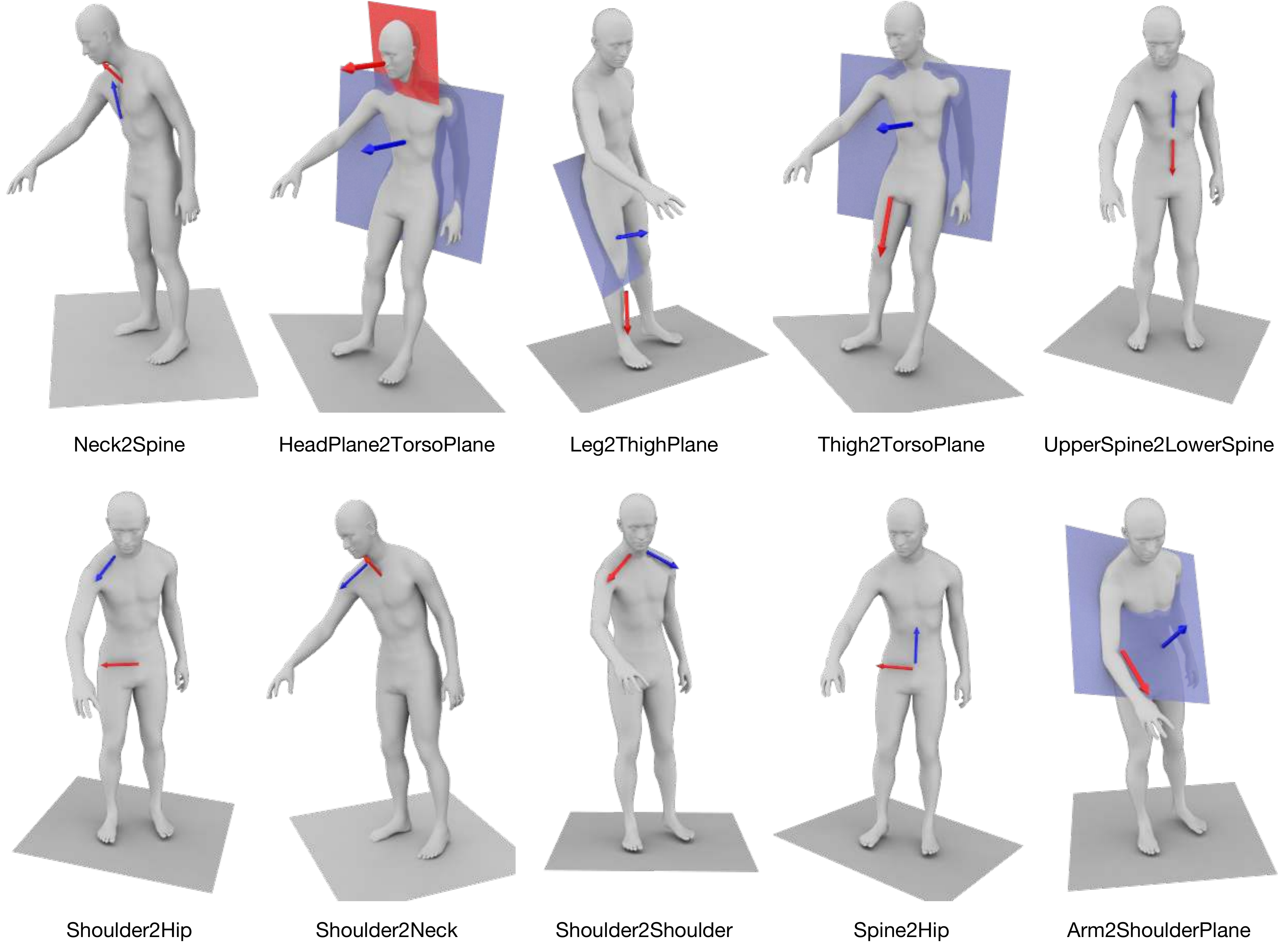}
		\end{tabular}
	\end{center}
	\vspace{-0.8cm}
	\captionof{figure}{{\bf Definition of additional angles.}\vspace{0.2cm}}\label{fig:angle_prior}
	\end{@twocolumnfalse}
}]

\begin{figure*}[!ht]
	\centering
	\includegraphics[width=0.85\linewidth]{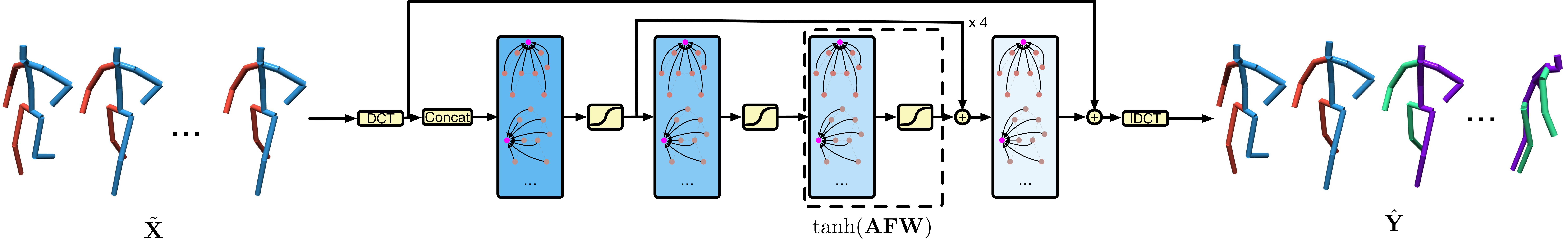}
	\caption{{\bf Overview of our generator.} Note that, here, we show a generator that predicts the whole body motion ($N=1$). In our experiments, however, we use the same architecture to predict the motion of human body parts.}
	\label{fig:generator}
\end{figure*}

\section{Details of Our Model}
\subsection{Pose Prior}
Our pose prior aims to model the distribution of valid human poses. As the validity of a pose mostly depends on the kinematics of its joint angles, 
instead of modeling the distribution of 3D joint coordinates, which couple the limb directions with their lengths, we propose to learn the distribution of limb directions only. In particular, given the $i$-th joint coordinate $\mathbf{J}_{i}\in\mathbb{R}^{3}$ and the coordinates of its parent joint $\mathbf{J}_{p_i}$, the limb direction can then be computed as
\begin{align}
	\mathbf{d}_i = \frac{\mathbf{J}_{i}-\mathbf{J}_{p_i}}{\|\mathbf{J}_{i}-\mathbf{J}_{p_i}\|_2}\;.
\end{align}
We then represent a human pose as the directions of all limbs $\mathbf{d} = [\mathbf{d}_1^T,\mathbf{d}_2^T,\cdots, \mathbf{d}_J^T]^T\in\mathbb{R}^{3J}$, where $J$ is the number of limbs, which, in our case, is equal to the number of joints. 

As mentioned in the main paper, we choose a simple network with 3 fully-connected layers to model our pose prior. To ensure the  
invertiblity of the network, we formulate each fully-connected layer as
\begin{align}
	\mathbf{f}^{'} = \sigma(\mathbf{f}\mathbf{Q}\mathbf{R} + \mathbf{b})\;,
\end{align}
where $\mathbf{f}^{'}\in\mathbb{R}^{3J}$ and $\mathbf{f}\in\mathbb{R}^{3J}$ are the feature vectors of the output and input, respectively; $\mathbf{Q}\in\mathbb{R}^{3J\times 3J}$ is an orthogonal matrix; $\mathbf{R}\in\mathbb{R}^{3J\times 3J}$ is an upper triangular matrix with positive diagonal elements; $\mathbf{b}\in\mathbb{R}^{3J}$ is the bias; $\sigma(\cdot)$ is the PReLU function.

\subsection{Angle Loss}
In Fig.~\ref{fig:angle_prior}, we visualise additional angles used to defined our kinematics constraints. In Table~\ref{tab:angle_prior}, we provide the
valid motion range for most angles used.

\begin{table}[!ht]
	\centering
	\resizebox{\linewidth}{!}{
		\begin{tabular}{@{\hskip 0mm}cccccc@{\hskip 0mm}}
			\toprule
			\multirow{2}{*}{Angles (in Degree)} & \multicolumn{2}{c}{Human3.6M~\cite{h36m_pami}} & & \multicolumn{2}{c}{HumanEva-I~\cite{sigal2010humaneva}}\\
			\cmidrule{2-3} \cmidrule{5-6}
			& LowerBound & UpperBound & & LowerBound & UpperBound \\\midrule
			Neck2Spine            & 0          & 124 &       & 0          & 50         \\
			HeadPlane2TorsoPlane  & 0          & 120  &      & -          & -          \\
			Leg2ThighPlane        & 80         & 180  &     & 72         & 166        \\
			Thigh2TorsoPlane      & 0          & 140  &      & 0          & 58         \\
			UpperSpine2LowerSpine & 110        & 180   &     & -          & -          \\
			Shoulder2Hip          & 0          & 84     &    & -          & -          \\
			Shoulder2Neck         & 32         & 134     &   & -          & -          \\
			Shoulder2Shoulder     & 83         & 180      &  & 128        & 180        \\
			Spine2Hip             & 60         & 120      &  & -          & -          \\
			Arm2ShoulderPlane     & -          & -        &  & 0          & 91         \\
			\bottomrule
		\end{tabular}
	}
	\caption{\textbf{Ranges of different angles.} Depending on the skeleton model, some angles are undefined in some dataset.}
	\label{tab:angle_prior}
\end{table}
\begin{figure}[!ht]
	\centering
	\includegraphics[width=0.4\linewidth]{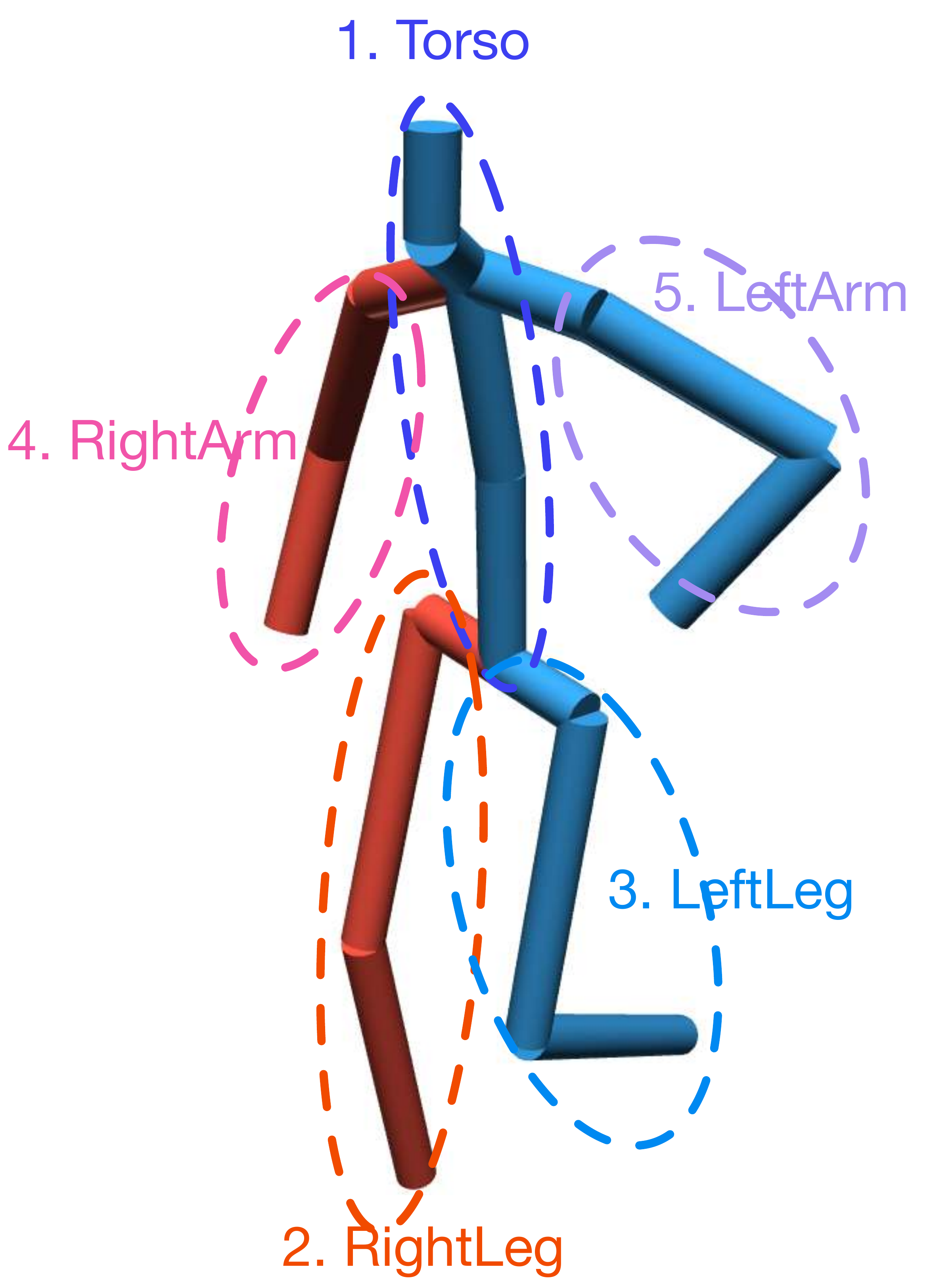}
	\caption{{\bf Body parts.} We divide a pose into 5 parts: 1. torso, 2. right leg, 3. left leg, 4. right arm and 5. left arm.}
	\label{fig:parts_n5}
\end{figure}
\subsection{Generator}
Recall that the input to the generator is $\Tilde{\mathbf{X}}=[\mathbf{x}_1,\mathbf{x}_2,\cdots,\mathbf{x}_H,\mathbf{x}_H,\cdots,\mathbf{x}_H]$ of length $H+T$. The first $H$ frames are the motion history, and the remaining $T$ frames are replications of last observed frame $\mathbf{x}_H$. The goal of the generator is to predict the DCT coefficients of the future motion $\Hat{\mathbf{Y}}\in\mathbb{R}^{D\times(H+ T)}$
given those of the replicated motion sequence, which we translate to learning motion residuals. Note that, to encourage a smooth transition between past poses and future ones, our generator not only predicts the future motion (corresponding to the last $T$ frames of $\Hat{\mathbf{Y}}$), but also recovers the past motion (corresponding to the first $H$ frames of $\Hat{\mathbf{Y}}$). Therefore, we define a reconstruction loss on the past motion as
\begin{align}
	\mathcal{L}_{past} = \|\Hat{\mathbf{Y}}[1:H]-\Tilde{\mathbf{X}}[1:H]\|_2^2\;,
\end{align}
where $\Hat{\mathbf{Y}}[1:H]$ and $\Tilde{\mathbf{X}}[1:H]$ are the first $H$ frames of $\Hat{\mathbf{Y}}$ and $\Tilde{\mathbf{X}}$, respectively. Furthermore, to enforce the limb length of the past poses to be the same as those of the future poses, we include the limb length loss
\begin{align}
	\mathcal{L}_{limb} = \sum_{t=1}^{T}\sum_{i=1}^{J}\|\Hat{l}_{t,i}-l_i\|_2^2\;,
\end{align}
where $\Hat{l}_{t,i}$ is the $i$-th limb length of $t$-th future pose, $l_i$ is the ground truth of $i$-th limb length obtained from the history poses and $J$ is the number of limbs. 
The loss weights of $(\mathcal{L}_{past},\mathcal{L}_{limb})$ for Human3.6M and HumanEva-I are both $(100,500)$.

The detailed architecture of our generator is shown in Fig.~\ref{fig:generator}. It consists of 4 residual blocks. Each block comprises 2 graph convolutional layers. It also contains two additional layers, one at the beginning, to bring the input to feature space, and the other at the end, to decode the feature to the residuals of the DCT coefficients.

\begin{figure*}[!ht]
	\centering
	\includegraphics[width=\linewidth]{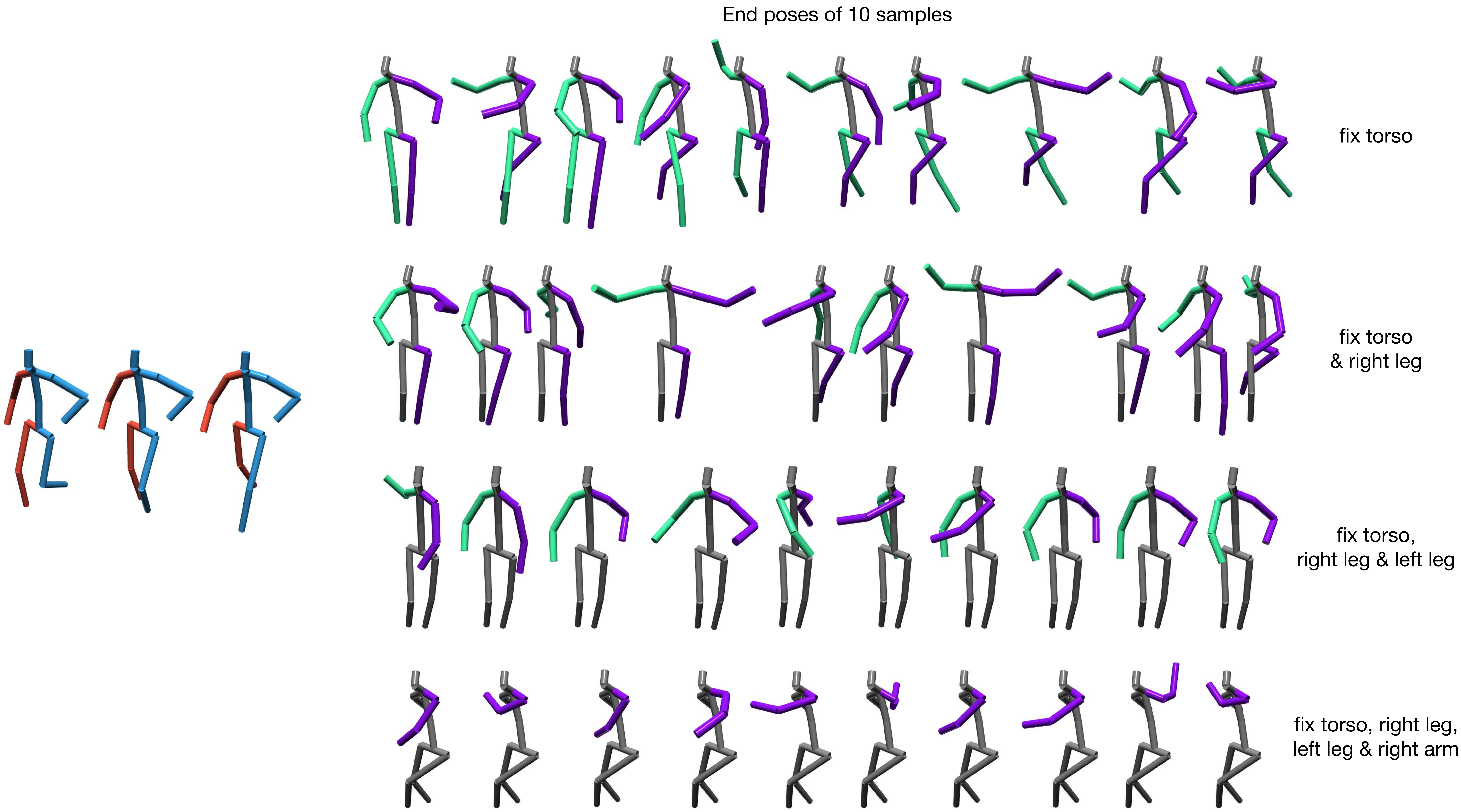}
	\caption{{\bf Results of controllable motion prediction with $N=5$.} In each row, we show the end poses of 10 samples predicted with the same motion for certain body parts. The controlled body parts are shown in gray.}
	\label{fig:control_pred_n5}
\end{figure*}
\begin{figure*}[!ht]
\centering
\includegraphics[width=\linewidth]{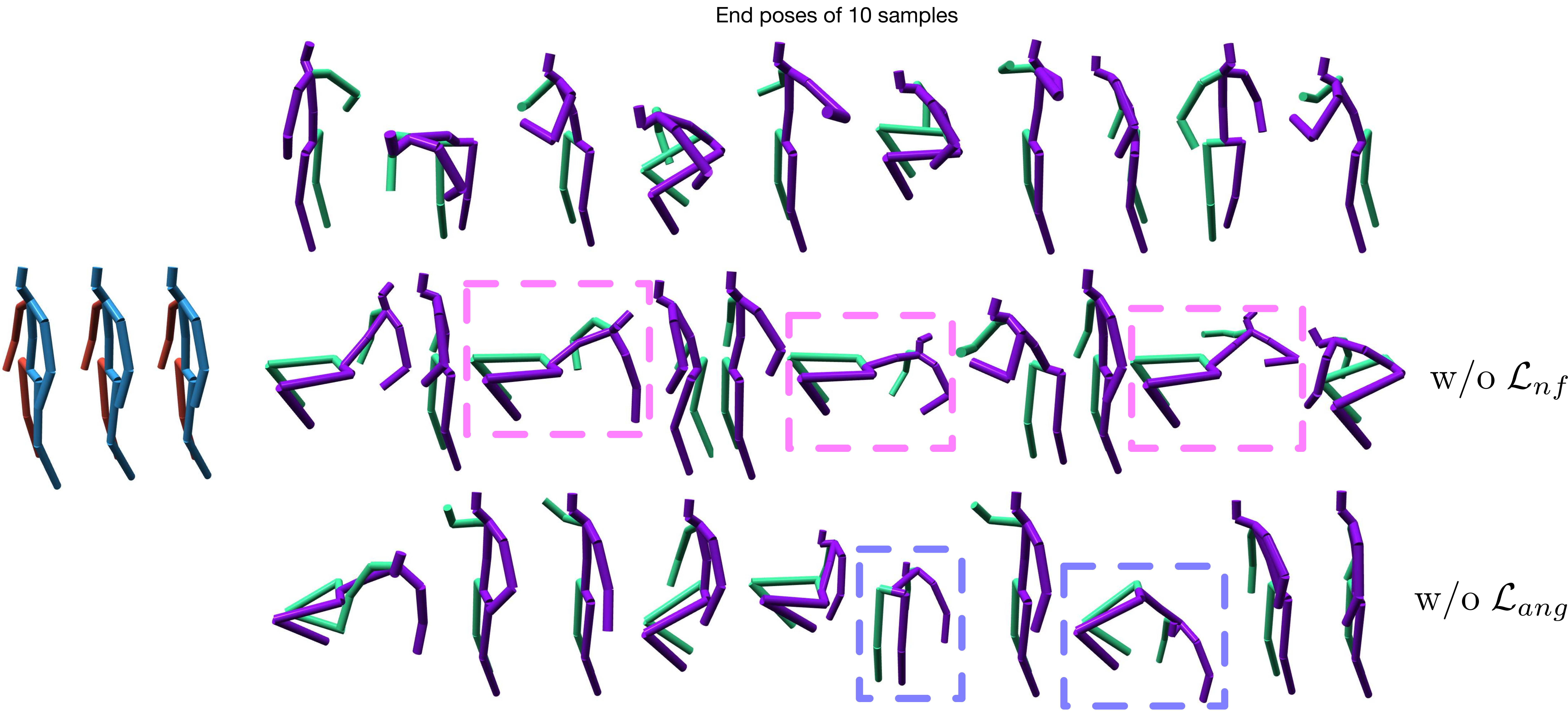}
\caption{{\bf Qualitative results of ablation study.} From top to bottom, we show the end pose of 10 samples of our model with all proposed losses, without the pose prior loss $\mathcal{L}_{nf}$ and without the angle loss $\mathcal{L}_{ang}$. Without the pose prior, the model predicts unlikely poses, as highlighted by the magenta boxes. When our model is trained without the angle loss, it produces invalid poses, highlighted by blue boxes.}
\label{fig:ablation_more}
\end{figure*}

\section{Implementation Details}
We implemented our network using Pytorch~\cite{paszke2017automatic} and used ADAM~\cite{kingma2014adam} to train it. The learning rate was set to 0.001 with a decay rate defined by the function
\begin{align}
	lr\_decay = 1.0 - \frac{\max(0,e - 100)}{400}\;,
\end{align}
where $e$ is the current epoch. 

For Human3.6M, our model observes the past 25 frames to predict the future 100 frames, and we use the first 20 DCT coefficients. For HumanEva-I, our model predicts the future 60 frames given the past 15 frames, and we use the first 8 DCT coefficients.

To generate the pseudo ground truth for the multi-modal reconstruction error ($\mathcal{L}_{mm}$), we follow the official DLow~\cite{yuan2020dlow}
implementation of the multi-modal version of FDE (MMFDE) and use the distance between
the last pose of the history to choose the pseudo ground truth. That is, for a training sample,
any other training sequence with similar last pose in terms of  Euclidean distance is
chosen to be a possible future. We set the distance threshold to 0.5 for both
Human3.6M and HumanEva-I.

\section{Additional Qualitative Results}
\subsection{Controllable Motion Prediction with $N>2$}
In the main paper, we divided a human pose into 2 parts ($N=2$): lower body and upper body, and show the results of predicting motions with same lower body motion but diverse upper body motions. Here, we provide qualitative results with $N=5$. In particular, as shown in Fig.~\ref{fig:parts_n5}, we split a pose into: 1. torso, 2. right leg, 3. left leg, 4. right arm and 5. left arm, following this order for prediction. As shown in Fig.~\ref{fig:control_pred_n5}, our model is able to perform different levels of controllable motion prediction given the detailed body parts segmentation.

\subsection{Ablation Study}
We show a qualitative comparison on the results of our model without either the pose prior loss $\mathcal{L}_{nf}$ or the angle loss $\mathcal{L}_{ang}$ in Fig.~\ref{fig:ablation_more}. This clearly demonstrates the dramatic decrease in pose quality in both cases.

\end{document}